%% file: main.tex
\documentclass[10pt,twocolumn,letterpaper]{article}

\usepackage[pagenumbers]{almost_wacv} 

\input{preamble}

\definecolor{froggygreen}{rgb}{0.01,0.60,0.01}
\usepackage[pagebackref,breaklinks,colorlinks,allcolors=froggygreen]{hyperref}
\usepackage{xurl}

\title{Automatic Fine-grained Segmentation-assisted Report Generation}

\author{Frederic Jonske$^*$
\and
Constantin Seibold
\and
Osman Alperen Koraş
\and
Fin Bahnsen
\and
Marie Bauer
\and
Amin Dada
\and
Hamza Kalisch
\and
Anton Schily
\and
Jens Kleesiek\\
\and
Institute for AI in Medicine, University Medicine Essen\\
Girardetstraße 2, 45131 Essen, Germany\\
{\tt\small (Frederic.Jonske, OsmanAlperen.Koras, Fin.Bahnsen, Marie.Bauer, Amin.Dada, }\\{\tt\small Hamza.Kalisch, Jens.Kleesiek)}{\tt\small @uk-essen.de, }{\tt\small constantinseibold@gmail.com, }{\tt\small antonputnik@gmail.com}
}

\begin{document}
\maketitle
\input{paper_v2}
{
    \small
    \bibliographystyle{ieeenat_fullname}
    \bibliography{bibliography}
}
\input{supplementary_v2}
\end{document}

%% file: preamble.tex
\usepackage{graphicx}
\usepackage{amsmath}
\usepackage{amssymb}
\usepackage{booktabs, makecell, tabularx}
\newcolumntype{C}{>{\centering\arraybackslash}X}
\usepackage{tcolorbox}

%% file: paper_v2.tex
\begin{abstract}
    Reliable end-to-end clinical report generation has been a longstanding goal of medical ML research. The end goal for this process is to alleviate radiologists' workloads and provide second opinions to clinicians or patients. Thus, a necessary prerequisite for report generation models is a strong general performance and some type of innate grounding capability, to convince clinicians or patients of the veracity of the generated reports. In this paper, we present ASaRG (\textbf{A}utomatic \textbf{S}egmentation-\textbf{a}ssisted \textbf{R}eport \textbf{G}eneration), an extension of the popular LLaVA architecture that aims to tackle both of these problems. ASaRG proposes to fuse intermediate features and fine-grained segmentation maps created by specialist radiological models into LLaVA's multi-modal projection layer via simple concatenation. With a small number of added parameters, our approach achieves a +0.89\% performance gain ($p=0.012$) in CE F1 score compared to the LLaVA baseline when using only intermediate features, and +2.77\% performance gain ($p<0.001$) when adding a combination of intermediate features and fine-grained segmentation maps. Compared with COMG and ORID, two other report generation methods that utilize segmentations, the performance gain amounts to 6.98\% and 6.28\% in F1 score, respectively. ASaRG is not mutually exclusive with other changes made to the LLaVA architecture, potentially allowing our method to be combined with other advances in the field. Finally, the use of an arbitrary number of segmentations as part of the input allows tracing elements of the report to the corresponding segmentation maps and verifying the groundedness of assessments.
\end{abstract}

\section{Introduction}
\label{sec: Introduction}

\begin{figure}[t]
    \centering
    \includegraphics[width=\linewidth]{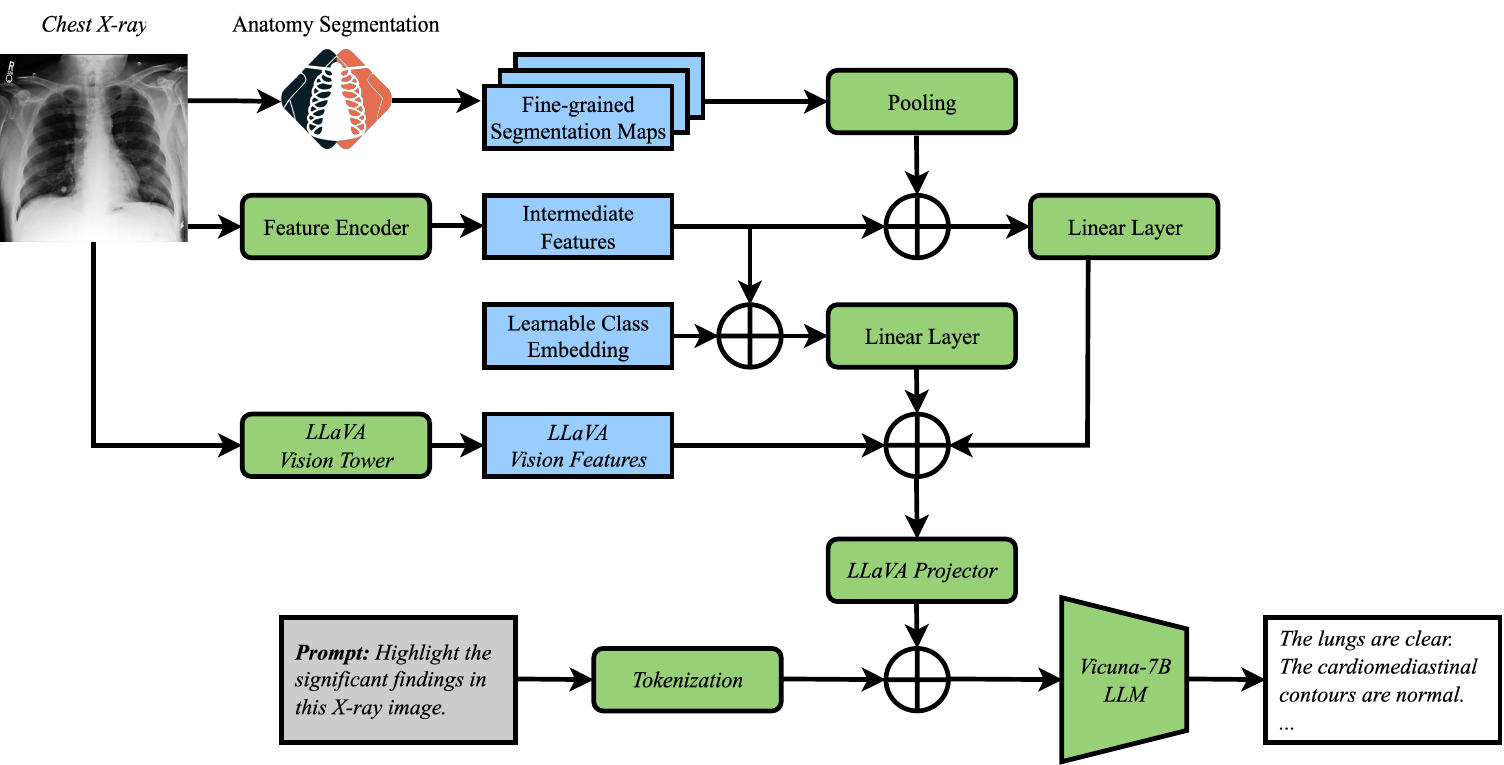}
    \caption{\textbf{The ASaRG architecture} - Model elements of ASaRG are highlighted in green and different input modalities are highlighted in blue. Plus symbols denote concatenation operations. Italics in any component denote that the component is part of original LLaVA architecture.}
    \label{fig: Visual Abstract}
\end{figure}

In recent years, multi-modal radiological report generation has made significant strides \cite{WANG2025103627}, both in terms of performance and supported modalities (e.g. \cite{Ham_CT2Rep_MICCAI2024, lei_autorg-brain_2024}), with generated reports slowly approaching the realm of human performance and already being sometimes preferable to human reports \cite{tanno_collaboration_2025}. The widespread interest in this field of research has two key reasons. For one, AI-driven report generation harbors immense potential for lightening the workload of radiologists in evaluating the image and creating the reports, as well as in explaining said report to patients without requiring the presence of clinicians. On the other hand, a strong report generation model can offer a potentially valuable second opinion in any case where a second opinion by another radiologist may not be readily available.

However, ML models that interact meaningfully with clinicians or patients do not only require (near-)human performance, but also a level of explainability or explicit grounding capability before they can be trusted with any responsibility. While recent work has increasingly emphasized these aspects, such grounding capability often comes at the cost of complex, purpose-built architectures \cite{tanida_interactive_2023, gu_complex_2024} that need to reinvent the wheel in many respects.

In this work, we present ASaRG, \textbf{A}utomatic \textbf{S}egmentation-\textbf{a}ssisted \textbf{R}eport \textbf{G}eneration. ASaRG proposes to tackle both the performance and grounding challenges by leveraging domain-specific visual features and fine-grained segmentation maps as additional inputs and extending the popular LLaVA architecture \cite{liu_visual_2023} to utilize these new inputs. The segmentation masks provide the report generation model with local-level cues about anatomical and pathological details and enable the grounding of report sections in the related segmentation masks. The domain-specific visual features provide additional global-level information that is complementary to features from LLaVA's vision encoder. The additional inputs are provided by two specialist medical models; LVM-Med \cite{m_h_nguyen_lvm-med_2023}, which provides intermediate visual embeddings, and an extended version of the CXAS framework \cite{seibold_accurate_2023}, which provides 212 full-size anatomical, pathological, and foreign objects segmentation maps. A lightweight addition to the original LLaVA projection layer aligns the additional modalities with the regular vision embeddings, both in terms of input size and embedding space layout, before concatenating all embeddings and feeding the entire sequence into the original LLaVA projector, greatly increasing overall performance.

Our contributions are as follows: 1) We propose to enhance medical report generation with LLaVA by extending the LLM input with two additional modalities, intermediate features and extremely fine-grained segmentations created by specialized medical models. 2) We explore different strategies for optimally fusing these new modalities into the existing LLaVA architecture with minimal parameter overhead. 3) We evaluate our resulting method on MIMIC-CXR \cite{johnson_mimic-cxr_2019}, where it significantly outperforms baseline LLaVA, despite freezing both the vision tower and LLM backbone of LLaVA compared to said baseline. ASaRG also beats competitive models that use smaller numbers of segmentation maps in Clinical Efficacy (CE) metrics. 4) With the explicit introduction of segmentation maps into the LLaVA model input, ASaRG also lays an easily extensible foundation for future research into grounded report generation. Our code will be made publicly available on publication.

\section{Related Works}
\label{sec: Related Works}

\subsection{Medical Report Generation}
\label{subsec: Medical Report Generation}

A number of recent publications have advanced the state of the art of medical report generation, influencing this work:

Li et al. extended the LLaVA framework using biomedical figure-caption pairs. The resuling LLaVA-med model \cite{li_llava-med_2023} outperforms state-of-the-art supervised approaches on three biomedical VQA datasets.

The MAIRA series of report generators \cite{bannur_maira-2_2024, hyland_maira-1_2024} innovated on the original LLaVA architecture with several minimal but highly influential changes. MAIRA-1 improved on previous report generators by extending the MIMIC-CXR dataset with GPT-paraphrased \cite{NEURIPS2020_1457c0d6} versions of all image-report pairs and choosing a CXR-specific image encoder, RAD-DINO \cite{perez-garcia_exploring_2025}. MAIRA-2 \cite{bannur_maira-2_2024} further built on this success by optionally incorporating multiple image views during generation. They further established a sentence-level factual correctness and grounding check as a novel report generation task.

Zhou et al. presented MedVersa \cite{zhou_medversa_2025}, a generalist multi-modal learner, as well as a 13 million annotations-strong multi-modal medical image-text benchmark. MedVersa outperformed competitors on this benchmark, in some instances by as much as 10\% compared to specialist models. 

Tu et al. created Med-PaLM M \cite{tu_towards_2024}, a multi-modal generalist biomedical AI capable of report generation, biomedical question answering, and image interpretation. Med-PaLM M was tested on an author-developed biomedical benchmark named MultiMedBench, demonstrating strong performance, and achieving a pairwise preference for AI reports of above 40\% when compared to those of clinicians. 

Rao et al. developed ReXErr-v1 \cite{rao_rexerr-v1_nodate}, a modernized report generation dataset, injecting typical human and AI errors into reports drawn from MIMIC-CXR image-report pairs, allowing future models to be trained with additional robustness against making the same errors.

Finally, report generation models have also made first steps into user studies recently. Tanno et al. created a report generation and conversation framework called Flamingo-CXR \cite{tanno_collaboration_2025}, which they tested by pitting generated reports against human clinician reports. They found that automatically generated reports were often equally or more preferred by human raters, demonstrating the value of further research into clinical report generation.

\subsection{Region-based Methods}
\label{subsec: Region-based Methods}

Region-based approaches have been a staple of image captioning and grounded VQA tasks outside of medical report generation for some time now \cite{anderson2018bottom}. Modern approaches typically leverage pre-existing LLMs and utilize either bounding boxes \cite{zhang2024gpt4roi, wang2023all, wang2025marten}, or pixel-level segmentations \cite{guo2024regiongpt, rasheed2024glamm} to identify salient image regions. It has been shown that these regions of interest (RoI) can be captured automatically \cite{guo2024regiongpt, zhang2024gpt4roi}, although the use of specialized instruction datasets is also common \cite{rasheed2024glamm, wang2023all, wang2025marten}. Such approaches are known to generally improve the question answering and reasoning capabilities of the incorporated LLMs \cite{guo2024regiongpt, rasheed2024glamm, wang2023all, wang2023visionllm, wang2025marten, zhang2024gpt4roi}, but can also lead to gains across other tasks, such as image captioning, object detection, or classification \cite{guo2024regiongpt, rasheed2024glamm, wang2023all}. Reference to specific image RoIs has also proven effective in reducing model hallucinations \cite{wang2025marten}.

Recognizing these inherent advantages, a small number of medical report generation publications have adopted region-based approaches.

With the explicit goal of naturally grounding generated reports in the source image, Tanida et al. developed the Radiology-guided Report Generation (RGRG) algorithm \cite{tanida_interactive_2023}. RGRG generates bounding box-based ROI suggestions and aggregates information from these ROIs into one final report, achieving strong performance compared to contemporary models.

Gu et al. developed a report generation framework titled Complex Organ-mask-Guided Report Generation (COMG) \cite{gu_complex_2024}, which combines information from four anatomical region segmentations with image and text embeddings using auxiliary encoders and cross-attention. Gu et al. further developed the Organ-Regional Information Driven (ORID) framework \cite{gu_orid_2025}, which draws on five region segmentations in addition to language and image inputs.

ASaRG differentiates itself from COMG and ORID by incorporating significantly more fine-grained segmentations, while crafting a less complex and more clinically performant extension of the LLaVA architecture.

\section{Methods}
\label{sec: Methods}

\subsection{Dataset Acquisition}
\label{subsec: Dataset Acquisition}

All experiments in this chapter are performed on the MIMIC-CXR dataset \cite{johnson_mimic-cxr_2019}\footnote{provided by Physionet \cite{Goldberger2000-vl} after obtaining permission}. This dataset contains 377'110 de-identified images in 227'835 radiographic studies from the Beth Israel Deaconess Medical Center in Boston, Massachusetts, United States, and represents the largest publicly available chest X-ray report generation dataset. The reports were converted to a VQA format using an LLM finetuned on clinical data, integrating the input image into a generated question prompt, and retaining the findings section of the report as the target answer. A detailed accounting of this procedure can be found in the supplementary.

The implicit assumption during finetuning of the LLaVA model (cf. Sect. \ref{subsec: Report Generation with LLaVa}) is that LLaVA's vision tower captures all relevant visual information and provides it to the LLM stage in the form of aligned embeddings. Apart from the fact that this cannot be fully true, purely because LLaVA is still far away from human expert performance, there are good reasons to assume that additional, domain-specific information may help capturing the essence of analyzed X-rays \cite{jonske_why_2025, hyland_maira-1_2024, gu_complex_2024}. Thus, in addition to the radiologist report, the following data is aggregated during dataset acquisition: Firstly, with the intuition that segmentation models are capable of decoding segmentations from an intermediate latent space representation of sufficient size, such representations are extracted for all images using the LVM-Med segmentation foundation model \cite{m_h_nguyen_lvm-med_2023}. Features are extracted from the final layer before the output head. Secondly, full-size, fine-grained segmentation maps are created by CXAS \cite{seibold_accurate_2023} with the intuition that more information may prove more useful at the cost of additional compute overhead. The original 158 classes in CXAS are extended by 54 additional classes, which include pathologies and foreign objects. Their segmentation maps are generated by additionally finetuning CXAS on ChestX-Det \cite{9395510} and the CLiP, Catheters and Line Positions datasets \cite{Tang2021-ge}. A summary of all segmentation classes can be found in the supplementary.

\subsection{Report Generation with LLaVA}
\label{subsec: Report Generation with LLaVa}

In this section, we briefly recall how the LLaVA architecture works, so that incremental improvements can be understood more easily. The \textbf{L}arge \textbf{La}nguage \textbf{a}nd \textbf{V}ision \textbf{A}ssistant (LLaVA) \cite{liu_visual_2023} is a groundbreaking advancement in multi-modal, image-text model research, with a great variety of models based on it being published in recent years \cite{bannur_maira-2_2024, hyland_maira-1_2024, liu_improved_2024, liu_llava-next_2024, li_llava-med_2023}. LLaVA is extremely simple and powerful, consisting of only an image encoder, a multi-modal projector layer, and an LLM, making LLaVA modular and allowing it to scale with later model developments (such as the recent LLama-3 \cite{dubey_llama_2024} or Qwen-1.5 \cite{bai_qwen_2023} integrations). 

A forward pass in LLaVA is a four-step process. Firstly, the input image $I$ is encoded using the image encoder $V$, which is typically a CLIP-pretrained \cite{radford_learning_2021} ViT \cite{dosovitskiy_image_2021} architecture, although it has been successfully exchanged for purpose-built vision encoders in literature \cite{bannur_maira-2_2024, hyland_maira-1_2024}. The encoded image features $\mathcal{F}_I = VE(I)$ are passed through a projector layer $P$, aligning the image features with the embedding space of the language model. Parallel to this, text embeddings $\mathcal{F}_T=TE(T)$ are created from the instruction prompt $T$ via tokenization and embedding $TE$. Finally, both embedding vectors are concatenated and fed into the LLM, which produces the desired output:
\begin{equation}
    O=LLM(CAT(P(\mathcal{F}_I), \mathcal{F}_T)).
    \label{eq1}
\end{equation}

LLaVA was originally trained in two stages. During stage one, the parameters of the vision and language models are frozen, and only the multi-modal projector is trained, forcing it to learn a projection from the vision tower's latent space to that of the LLM. In the second stage, the LLM parameters are trainable as well, improving the model's overall ability to reason from the added visual information. We specifically adopt LLaVA based on Vicuna-7B \cite{zheng_judging_2023} and a ViT L/14 vision tower (336x336 resolution) \cite{dosovitskiy_image_2021}. Since the model has been trained using conversation prompts, it can effectively be used out-of-the-box for medical report generation by treating it as a single-turn conversation. 

\subsection{ASaRG}
\label{ASaRG}

ASaRG extends LLaVA by modifying the existing projector layer $P$ to include additional features $\mathcal{F}_{new}$, such that:
\begin{equation}
    O=LLM(CAT(P^*(\mathcal{F}_I, \mathcal{F}_{new}), \mathcal{F}_T)).
    \label{eq2}
\end{equation}
The modified projector layer $P^*$ is defined as:
\begin{equation}
    P^*(\mathcal{F}_I, \mathcal{F}_{new})=P(g(\mathcal{F}_I, \mathcal{F}_{new})), with
    \label{eq3}
\end{equation}
\begin{equation}
    \mathcal{F}_{new}=f(C, R, S)
    \label{eq4}
\end{equation}
where $P$ is LLaVA's original projector layer, $C$ is a learnable class embedding, and $R$ and $S$ denote our extracted image features and fine-grained segmentation maps belonging to an image $I$. $f$ and $g$ are functions that allow information from their inputs to interact in some way. We will see later that an optimal function choice has a massive impact on performance. The following subsection details and motivates our candidate choices.

We begin by creating a learnable class embedding $C$
\begin{equation}
    C=ReLU(Conv1D(Emb(Range[1, 256], d), c)).
    \label{eq5a}
\end{equation}
of dimension $b * c=288 * d=1024$. In essence, this class embedding can be understood as "class labels", similarly to a positional encoding in a vision transformer \cite{dosovitskiy_image_2021}. These classes and what they represent are effectively arbitrary and can correspond to any patterns encoded in the extracted image features $R$. The extracted features $R$ are brought into the same shape: 
\begin{equation}
    R_{stack} = Stack(ReLU(Linear(R, 1024)), 288).
    \label{eq5b}
\end{equation}
The different methods we test diverge at this point:

\textbf{Feature Replacement} - $C$ and $R_{stack}$ are "mixed" via AdaptiveInstanceNorm \cite{huang_arbitrary_2017} and a 1D-convolution, with:
\begin{equation}
    R_I=Conv1D(AdaIN(C, R_{stack})).
    \label{eq5c}
\end{equation}
The result has the same shape ($b * 576 * 1024$) as the original LLaVA vision tower output and is fed into the original pretrained projection layer. The original vision tower output is not used. The rationale behind this approach is that the intermediate features should already contain visual information condensed from a specialist model designed for fine-grained segmentation.

\textbf{Learned Mixing} - $C$ and $R_{stack}$ are mixed by concatenating them along the embedding axis and feeding them into a linear layer $L_1$. Image features derived from the vision tower are mixed with the result using the same process:
\begin{equation}
    R_I=Lin_1(CAT(C, R_{stack})),
    \label{eq6}
\end{equation}
\begin{equation}
    g(\mathcal{F}_I, \mathcal{F}_{new}(R_I))=Lin_2(CAT(\mathcal{F}_I, R_I)).
    \label{eq7}
\end{equation}
This allows information from three sources - inherent bias, intermediate features from a specialist segmentation model, and features from a generalist vision model - to interact and complement one another.

\textbf{Weighted Addition} - Class embeddings and intermediate features are mixed as in Eq. \ref{eq6}. Interaction with the information from the vision tower is facilitated through weighted addition with a learnable weight $\alpha$:
\begin{equation}
    g(\mathcal{F}_I, \mathcal{F}_{new}(R_I))=\mathcal{F}_I + \alpha R_I,
    \label{eq8}
\end{equation}
The addition process has the benefit of fusing the new information into the projection layer output without significantly altering previous model weight "meanings", so long as $\alpha$ remains small.

\textbf{Concatenation} - Class embeddings and intermediate features are mixed as in Eq. \ref{eq6}. The result is concatenated to the vision tower's image features along the channel axis:
\begin{equation}
    g(\mathcal{F}_I, \mathcal{F}_{new}(R_I))=CAT(\mathcal{F}_I, R_I).
    \label{eq9}
\end{equation}
Any interaction between these information sources is restricted to the original projection layer and LLM. This has the advantage of requiring fewer parameters and fully conserving unaltered information from the vision tower, thus making the most effective use of LLaVA's pretraining.

To additionally include information from the fine-grained segmentations, the segmentation maps $S$ are first pooled via AdaptiveAveragePooling and a 1D-convolution. Similarly to how elements with different receptive fields can interact to combine local and global information in SWin-UNeTr \cite{hatamizadeh2021swin}, we allow the global features and our pooled segmentation maps $S_{loc.}$ to interact via a linear layer:
\begin{equation}
    S_I=Lin(CAT(R_I, S_{loc.})),
    \label{eq10a}
\end{equation}
\begin{equation}
     S_{loc.}=Conv1D(AAP(S)).
    \label{eq10b}
\end{equation}
Information from the fine-grained segmentations $S_I$ is fused into the modified projector in the same way that our extracted image features $R_I$ previously were. As later experiments confirm, concatenation is the preferable fusion method, yielding: 
\begin{equation}
    g(\mathcal{F}_I, \mathcal{F}_{new}(R_I, S_I))=CAT(\mathcal{F}_I, R_I, S_I)
    \label{eq11}
\end{equation}
The entire process is visualized in Fig. \ref{fig: Visual Abstract}.

\section{Experimental Setup}
\label{Experimental Setup}

\subsection{Experiments}
\label{subsec: Experiments}

\subsubsection{Finding the Optimal Fusion Method}

We test each fusion function described above by finetuning LLaVA models with the modified projector layer on our VQA-style MIMIC dataset. For weighted addition, the weighting parameter is initialized once at $\alpha=0$ and $\alpha=1$. $\alpha=0$ is chosen to allow the model to slowly adjust weights and increase reliance on new information over time, without destroying weight configurations acquired during pretraining, while $\alpha=1$ is chosen to explore immediately forcing the model to use newly given information and to prevent $\alpha$ permanently staying at zero. 
The training time for all experiments is limited to 1 epoch to minimize the computational overhead. While this concession is somewhat suboptimal because performance only saturates after at least 3 \cite{hyland_maira-1_2024} to at most 30 epochs \cite{gu_orid_2025}, depending on the experiment or literature comparison, probing experiments showed that performance gaps could usually already be observed quite clearly after one or two epochs. All other hyperparameters can be found in the supplementary.

\subsubsection{Adding Fine-grained Segmentation Maps}

After an optimal fusion method is identified, we add segmentation maps to ASaRG by extending the projector layer as described in Sect. \ref{ASaRG}. As a total of 212 segmentation maps add a significant computational and (V)RAM overhead, adding the segmentation maps involves modifying the training recipe to a two-stage process for this set of experiments. First, we train for one epoch with all parameters being trainable. Afterwards, all elements of the modified projector layer that handle segmentation inputs are added and randomly initialized. We then finetune for a second epoch, keeping only the parameters of the projector layer trainable to counteract the compute and VRAM overhead. We test both a features-only and a features+segmentations approach in this two-stage scenario, and compare against a two-stage-trained LLaVA and a fully trainable LLaVA that was trained for two epochs.
In addition to testing the use of maximally fine-grained segmentations, a superclasses setup is explored, where segmentation maps are aggregated into one of the 18 anatomical CXAS superclasses, a pathology superclass, or a foreign objects superclass via boolean addition. This significantly reduces the computational cost and may alter the performance in either direction, as detailed maps are taken away, but segmentation noise is partially removed during aggregation.

\subsection{Reported Metrics}
\label{subsec: Reported Metrics}

In the interest of comparability and a more comprehensive assessment, we report a broad collection of performance metrics. Among lexical metrics, we report BLEU-scores \cite{papineni_bleu_2001} based on length-1 and -4 n-grams, ROUGE-L (longest common subsequence) \cite{lin_rouge_2004}, CIDEr-D \cite{vedantam_cider_2015}, and METEOR \cite{lavie_meteor_2007} scores. For semantic validity, Clinical Efficacy (CE) scores are reported, based on the standard CheXbert \cite{smit_combining_2020}. Reported CES values refer to the full 14-class micro-averaged F1 score, recall, and precision. Uncertainties are reported for all experiments by way of repeating the entire experiment (finetuning, evaluation, scoring) 4 times, each time from different random initializations of the non-pretrained parts of the ASaRG architecture. Note that this measures training stability, but misses possible data-driven uncertainty, meaning reported values somewhat underestimate true uncertainty.

Reported p-values are derived for two groups of results with a Welch's t-test \cite{welch_generalization_1947}, based on CE F1 score. Since, as will be shown later, the preconceived expectation of performance improvements gained by ASaRG is reasonable, a one-sided test is applied. The resultant p-value effectively states the probability that any observed performance gain of some method A compared to another method or baseline B is due to random chance.

It should be noted here that common report generation metrics, especially lexical ones, are well-known to be gameable to some degree \cite{boag_baselines_2020, vedantam_cider_2015} and that they can suffer from both false positives and false negatives. The most glaring example of this is the antagonistic report. If a report is copied and a singular negation added to invert the key diagnosis, a patient may inadvertently be exposed to harm. A lexical metric such as BLEU would give such a report a near-perfect score, whereas a semantic metric would correctly indicate it as bad. A general solution to this issue remains an open research question in the field at the time of writing. Consequently, we interpret result comparisons primarily via the clinically more relevant Clinical Efficacy Score. However, for most comparisons, the same trend is demonstrated across most or all metrics.

\section{Results}
\label{sec: Results}

\begin{table*}[t]
    \centering
    \begin{tabularx}{\textwidth}{|C|C|C|C|C|C|C|C|C|}
        \hline
        \textbf{Method} & \textbf{BLEU-1} & \textbf{BLEU-4} & \textbf{METEOR} & \textbf{ROUGEL} & \textbf{CIDEr-D} & \textbf{CE (Pr)} & \textbf{CE (Rc)} & \textbf{CE (F1)} \\
        \hline
        LLaVA (Baseline) & 0.2126 \mbox{± }0.0012  & 0.0539 \mbox{± }0.0014 & \textbf{0.1668 \mbox{± }0.0014} & 0.1937 \mbox{± }0.0015 & 0.2573 \mbox{± }0.0082 & 0.4665 \mbox{± }0.0029 & 0.3179 \mbox{± }0.0056 & 0.3781 \mbox{± }0.0046 \\
        \hline
        +Features, L. Mixing & 0.2056 \mbox{± }0.0058 & 0.0504 \mbox{± }0.0028 & 0.1602 \mbox{± }0.0017 & 0.1881 \mbox{± }0.0024 & 0.2442 \mbox{± }0.0203 & 0.4375 \mbox{± }0.0117 & 0.2793 \mbox{± }0.0178 & 0.3409 \mbox{± }0.0168 \\
        \hline
        +Features, Addition, $\alpha_{init}=0$ & 0.2134 \mbox{± }0.0010 & 0.0547 \mbox{± }0.0006 & 0.1661 \mbox{± }0.0006 & 0.1935 \mbox{± }0.0004 & 0.2583 \mbox{± }0.0052 & 0.4631 \mbox{± }0.0034 & 0.3117 \mbox{± }0.0012 & 0.3726 \mbox{± }0.0015 \\
        \hline
        +Features, Addition, $\alpha_{init}=1$ & 0.2151 \mbox{± }0.0010 & 0.0541 \mbox{± }0.0007 & 0.1640 \mbox{± }0.0006 & 0.1925 \mbox{± }0.0011 & 0.2623 \mbox{± }0.0037 & 0.4656 \mbox{± }0.0021 & 0.3251 \mbox{± }0.0026 & 0.3828 \mbox{± }0.0020 \\
        \hline
        +Features, Concat. & \textbf{0.2195 \mbox{± }0.0019} & \textbf{0.0569 \mbox{± }0.0007} & 0.1664 \mbox{± }0.0007 & \textbf{0.1955 \mbox{± }0.0007} & \textbf{0.2798 \mbox{± }0.0064} & \textbf{0.4691 \mbox{± }0.0049} & \textbf{0.3291 \mbox{± }0.0013} & \textbf{0.3868 \mbox{± }0.0016} \\
        \hline
    \end{tabularx}
    \caption{\textbf{Intermediate Feature Fusion}}
    \small
    Results for the tested configurations of ASaRG with intermediate features included. Evaluation is performed on the holdout test set after 1 epoch of finetuning. Uncertainties are derived by executing the entire experiment four times, including finetuning. Best results are in bold. Note that replacement of the original vision tower with the intermediate LVM-Med features does not converge to a meaningful solution.
    \label{tab: Intermediate Feature Fusion}
\end{table*}

\begin{table*}[t]
    \centering
    \begin{tabularx}{\textwidth}{|C|C|C|C|C|C|C|C|C|}
        \hline
        \textbf{Method} & \textbf{BLEU-1} & \textbf{BLEU-4} & \textbf{METEOR} & \textbf{ROUGEL} & \textbf{CIDEr-D} & \textbf{CE (Pr)} & \textbf{CE (Rc)} & \textbf{CE (F1)} \\
        \hline
        LLaVA, two-stage & 0.2103 \mbox{± }0.0010 & 0.0518 \mbox{± }0.0010 & 0.1672 \mbox{± }0.0007 & 0.1910 \mbox{± }0.0006 & 0.2441 \mbox{± }0.0033 & 0.4762 \mbox{± }0.0054 & 0.3262 \mbox{± }0.0030 & 0.3872 \mbox{± }0.0037 \\
        \hline
        LLaVA, fully ft'd & 0.2278 \mbox{± }0.0013 & 0.0561 \mbox{± }0.0010 & 0.1622 \mbox{± }0.0009 & 0.1928 \mbox{± }0.0012 & 0.2902 \mbox{± }0.0089 & 0.4803 \mbox{± }0.0013 & 0.3508 \mbox{± }0.0008 & 0.4055 \mbox{± }0.0009 \\
        \hline
        +Features, two-stage & 0.2154 \mbox{± }0.0003 & 0.0544 \mbox{± }0.0002 & 0.1677 \mbox{± }0.0006 & 0.1917 \mbox{± }0.0004 & 0.2516 \mbox{± }0.0070 & 0.4752 \mbox{± }0.0031 & 0.3396 \mbox{± }0.0037 & 0.3961 \mbox{± }0.0036 \\
        \hline
        +Features, +SegMaps, two-stage & 0.2301 \mbox{± }0.0007 & 0.0607 \mbox{± }0.0003 & 0.1711 \mbox{± }0.0008 & 0.2009 \mbox{± }0.0008 & 0.2901 \mbox{± }0.0042 & 0.4903 \mbox{± }0.0035 & \textbf{0.3596 \mbox{± }0.0036} & \textbf{0.4149 \mbox{± }0.0034} \\
        \hline
        +Features, +SegMaps, two-stage, SC-only & 0.2298 \mbox{± }0.0012 & 0.0612 \mbox{± }0.0008 & \textbf{0.1716 \mbox{± }0.0010} & 0.2011 \mbox{± }0.0010 & \textbf{0.2944 \mbox{± }0.0104} & \textbf{0.4905 \mbox{± }0.0019} & 0.3594 \mbox{± }0.0019 & 0.4148 \mbox{± }0.0015 \\
        \hline
        COMG$^\dagger$ \cite{gu_complex_2024} & 0.363 & \textbf{0.124} & 0.128 & \textbf{0.290} & * & 0.424 & 0.291 & 0.345 \\
        \hline
        ORID$^\dagger$ \cite{gu_orid_2025} & \textbf{0.386} & 0.117 & 0.150 & 0.284 & * & 0.435 & 0.295 & 0.352 \\
        \hline
    \end{tabularx}
    \caption{\textbf{Experiments with (fine-grained) Segmentation Maps}}
    \small
    Results for the tested configurations of ASaRG. Evaluation is performed on the holdout test set after 2 epochs of training. Two-stage denotes runs which were created by fully finetuning all available parameters for one epoch, and finetuning for a second epoch where only the parameters of the projector layer are trainable. A baseline LLaVA is also finetuned fully for two epochs (labelled 'fully ft'd'). Where segmentation maps are fused into the projector layer (+SegMaps), the newly added parameters are randomly initialized and trained during this epoch for the first time. Uncertainties are derived by executing the entire experiment four times, including finetuning. Daggers denote values reported in literature where segmentations are also used as part of the input. Best results are in bold.
    \label{tab: Experiments with (Fine-grained) Segmentation Maps}
\end{table*}

\begin{table*}[t]
    \centering
    \begin{tabularx}{\textwidth}{|C|C|C|C|C|C|C|C|C|}
        \hline
        \textbf{Method} & \textbf{BLEU-1} & \textbf{BLEU-4} & \textbf{METEOR} & \textbf{ROUGEL} & \textbf{CIDEr-D} & \textbf{CE (Pr)} & \textbf{CE (Rc)} & \textbf{CE (F1)} \\
        \hline
        LLaVA, fully ft'd & 0.2328 \mbox{± }0.0012 & 0.0530 \mbox{± }0.0005 & 0.1527 \mbox{± }0.0006 & 0.1841 \mbox{± }0.0004 & 0.2788 \mbox{± }0.0031 & 0.4544 \mbox{± }0.0026 & 0.3646 \mbox{± }0.0026 & 0.4046 \mbox{± }0.0025 \\
        \hline
        +Features, two-stage & 0.2414 \mbox{± }0.0005 & 0.0604 \mbox{± }0.0005 & 0.1594 \mbox{± }0.0007 & 0.1937 \mbox{± }0.0005 & 0.2982 \mbox{± }0.0021 & 0.4759 \mbox{± }0.0008 & 0.3802 \mbox{± }0.0020 & 0.4227 \mbox{± }0.0015 \\
        \hline
        +Features, +SegMaps, two-stage & \textbf{0.2518 \mbox{± }0.0015} & \textbf{0.0671 \mbox{± }0.0014} & \textbf{0.1654 \mbox{± }0.0010} & \textbf{0.2027 \mbox{± }0.0009} & \textbf{0.3337 \mbox{± }0.0111} & \textbf{0.4832 \mbox{± }0.0032} & \textbf{0.3981 \mbox{± }0.0026} & \textbf{0.4365 \mbox{± }0.0022} \\
        \hline
    \end{tabularx}
    \caption{\textbf{Experiments with longer training time}}
    \small
    Results for 5 epochs of training. ASaRG variants perform two-stage training as before, except with 2+3 instead of 1+1 epochs. Uncertainties are derived by training once and evaluating four times. While this dramatically lowers compute overhead, it inevitably underestimates the total uncertainty somewhat. Best results are in bold.
    \label{tab: Experiments with longer training time}
\end{table*}

The results for both sets of experiments can be found in Tab. \ref{tab: Intermediate Feature Fusion} and Tab. \ref{tab: Experiments with (Fine-grained) Segmentation Maps}, where they are also compared to baselines trained and evaluated with the same hyperparameters and dataset. We additionally report literature values from \cite{gu_complex_2024} and \cite{gu_orid_2025}, where a smaller number of segmentations have also been used for report generation. A class-wise breakdown of the performance for all model variants, based on the CE F1 Score, can be found in the supplementary. \\
Firstly, we note that both Replacement and fusion via Learned Mixing do not work well for our purposes, scoring significantly lower than the LLaVA baseline (no useful results and -3.73\% CE F1 score, respectively). In the case of Replacement, the apparent lesson is that the intermediate features alone either do not carry all necessary information to formulate an authentic report, or require significantly more time or additional parameters to successfully retrain the projector layer. Conversely, for Learned Mixing, the performance does appear to slowly amortize, after initially tanking, because the alignment of visual features and language model is no longer given due to the mixing. \\
Depending on the chosen value of $\alpha$, Weighted Addition scores below or above the baseline, with $\alpha=1$ offering increased performance (+0.47\%, $p=0.090$) despite also initially un-aligning vision and language features. We note that in all cases, $\alpha$ tended to remain near the initial value, with a standard deviation of around 0.01, despite the optimizer being principally capable of assigning a higher learning rate to the parameter, implying that Weighted Addition can principally work with any weighting. \\
Finally, at +0.87\% and $p=0.021$, Concatenation achieves the most significant performance boost, with the added benefit of maintaining any previous feature alignment. This improvement comes at the cost of only 0.06\% additional parameters, or 0.43\% if the parameters of LVM-Med are counted as well. We note that concatenation also comes with the advantage of allowing an effectively arbitrary number of additional information sources, such as features from additional specialist models, more segmentations, lab values, and so forth. This provides a simple and obvious starting point for future extension of this method. \\
Fusing the down-pooled segmentation maps into the LLM input delivers a significant additional performance boost of 1.88\% ($p<0.001$), or 2.77\% ($p<0.001$) compared to baseline LLaVA. Even compared to a baseline that performs full finetuning for two epochs - instead of the two-stage process of full finetuning and projector-only finetuning as ASaRG does to reduce compute overhead - ASaRG compares favorably, with a performance gain of 0.94\% ($p=0.007$). The parameter overhead of this improvement is +0.09\%, or 1.08\% if the parameters of LVM-Med and CXAS are both counted as well. \\
To verify that reported performance gains do not disappear in training scenarios with longer training time, we conduct a small probing experiment across 5 epochs. The results (cf. Tab. \ref{tab: Experiments with longer training time} strongly suggest that not only does ASaRG still offer performance gains at longer training times, but also that the gap widens considerably.
Interestingly, our performance gains are matched by a version of ASaRG that combines the original 212 segmentation maps into 20 superclasses, with a performance difference of <0.01\% between the two (A two-sided t-test confirms that the two approaches deliver the same performance with $p=0.989$). This implies that there is either a limit to the usefulness of extremely fine-grained segmentations in general, or the quality of the fine-grained segmentations still needs to improve to make them more useful in the future, and that superclass maps may be preferable when facing compute constraints. \\
Compared to literature, ASaRG variants systematically underperform in lexical metrics, but significantly overperform in semantic metrics, implying that they better understand the clinical implications of the analyzed X-rays, despite producing reports less similar to the original reports. To verify that the decreased lexical metric performance does not indicate unreadable or clinically impossible reports, we conducted a survey on 300 test set samples with the help of a clinician, which can be found in the supplementary.

\section{Discussion}
\label{Discussion}

\subsection{Interactions Between Features and Segmentation maps}
\label{subsec: Interactions Between Features and Segmentation maps}

To determine whether segmentation maps can be used for grounding, we randomized the order of the segmentation maps only for the test data and re-evaluated each run of the features+maps experiment. The result of this experiment can be found in Tab. \ref{tab: Experiments with (Fine-grained) Segmentation Maps}. The difference between this score and that for sorted segmentation maps represents the interpretable contribution that the segmentation maps bring to the table, which comes out to around +0.33\%. This implies that the remaining 0.61\% compared to the full finetuning baseline (or 2.44\% compared to a baseline trained with the two-stage method) can be attributed to one of the following: Firstly, the interaction between intermediate features and full segmentation maps, similar to how skip connections in a U-Net decoder add additional performance. Secondly, to some information inherent to the segmentation maps which are usefully "translated" by the projection layer, no matter which segmentation map they are in. One example of this could be a map that is heart-shaped but extremely large, pointing towards cardiomegaly, showing up in the "wrong" segmentation map in the shuffled scenario. We deem this scenario possible, albeit unlikely, as the number of such examples seems extremely limited to us. Finally, the remaining performance could be an artifact of assigning more parameters to the projector, which are influenced by the intermediate features. We posit that this is possible, but cannot explain the entire performance gain, because the inclusion of the segmentation maps and features only adds around 50\% more parameters to the projection layer than the inclusion of features alone does, but would have to explain two thirds of the total performance gain compared to the baseline. Further, a superclass-only ASaRG with fewer parameters delivers the same performance as ASaRG with all 212 classes included, implying that the effect of additional parameters has to be quite small.

\subsection{Grounding of Generated Reports}
\label{subsec: Grounding of Generated Reports}

\begin{figure}[t]
    \centering
    \includegraphics[width=\linewidth]{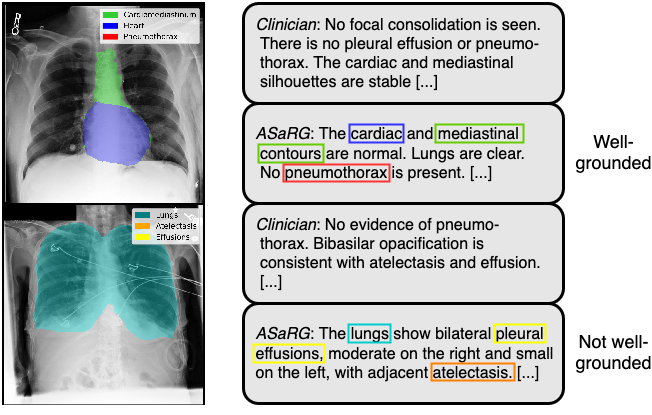}
    \caption{\textbf{Grounding with segmentation maps} - This graphic depicts two examples from the MIMIC-CXR test set with ground truths and generated reports. The colors highlight parts of the generated report and their corresponding segmentation maps. Correspondences are derived via NER Tagging and limited to a few example classes and reports are truncated for readability.}
    \label{fig: Segmap Grounding}
\end{figure}

ASaRG enables grounding of generated reports by way of checking findings in the generated report against available fine-grained segmentations for the associated image, access to which has improved the model performance. Examples of this process can be found in Fig. \ref{fig: Segmap Grounding}, which shows abbreviated ground truths and generated reports, highlighting both success and failure cases of our method. In the first example, the cardiomediastinum and heart are represented very well by their segmentation maps, implying that they are well-understood by the model and that the report is trustworthy in this respect. In the second example, atelectasis and pleural effusions are predicted by the generated report, in agreement with the original report. The lung segmentation also captures most of the opaque basilar lung areas. In contrast, however, atelectasis and pleural effusions are not predicted by the segmentation model. The latter would suggest that no pathology is present, while the former implies that the model has indeed understood that the lung volume in that area is mostly dysfunctional at the time of the X-ray. In sum, a researcher would now know to regard the generated report with more skepticism and may even draw conclusions regarding which part of their model they should improve - in this case, the pathology segmentation.

\subsection{Failure Modes}
\label{subsec: Failure Modes}

ASaRG and LLaVA exhibit several distinct failure modes which are discussed below, with some examples featured in the supplementary. Firstly, we note a strong propensity to reference non-existent prior studies, a hallucination well-known in literature \cite{hyland_maira-1_2024, bannur_maira-2_2024}. 
Secondly, generated reports often contain distances in cm, e.g. ETT to carina. While these are sometimes correct, the model has no exact length references and thus makes guesses based on learned numbers, which has also been reported previously \cite{hyland_maira-1_2024}. 
A general theme is the strong under-reporting of findings, compared to ground truths, with ASaRG/LLaVA false negatives (aggregated over all findings) averaging 13.6\%/13.8\%, compared to 6.7\%/6.2\% false positives. On average, reports contain 0.96/0.93 fewer pathologies per report than ground truth. This tendency correlates with the pathology count - ASaRG under-predicts in 69\% of cases if the ground truth has more than one, but only in 45\% of one-pathology cases. In some cases, this is caused by the one-view-one-report training scheme, because some ground truth reports derive findings only from specific views.
Finally, in rare cases, we note correct reports phrased so differently from the ground truth that even the evaluation by CE metric produces false negatives.

\subsection{Limitations}
\label{subsec: Limitations}

Several limitations apply to this work. Firstly, ASaRG is tested on only a single dataset, MIMIC-CXR. While this dataset has been the gold standard dataset for report generation on X-rays for some time, it leaves open the question whether our method would achieve similar success on other datasets or different modalities. 

Our VQA setting additionally introduces confounding factor when comparing to other methods. This may influence the general behavior of our models, although not the reported performance gains relative to baselines.

Another limitation lies in the limited training time, where we cannot fully dismiss the possibility of some tested approaches amortizing after a training time significantly longer than the one or two epochs in our experiments. 

We further note that while grounding can be established between segmentation maps and parts of generated reports easily, the limited performance gain compared to baseline implies limited grounding, although said gain and corresponding groundedness can be improved with future improvements to fine-grained segmentation approaches.

Finally, it is unclear whether all LLaVA offshoots will benefit from the inclusion of intermediate features or fine-grained segmentation maps to the same degree. It is possible that approaches such as MAIRA \cite{hyland_maira-1_2024} would observe smaller benefits than baseline LLaVA, because MAIRA's RAD-DINO-pretrained \cite{perez-garcia_exploring_2025} vision tower already encodes some part of the domain-specific information whose inclusion we credit with ASaRG's performance gains.

\section{Conclusion}
\label{sec: Conclusion}

In this paper, we presented ASaRG, a novel method for augmenting the performance of LLaVA-style medical report generation models using intermediate features and fine-grained segmentation maps generated by radiological specialist models. The concatenation-based fusion of the additional information sources offered meaningful performance gains across a wide range of performance metrics, even when comparing full finetuning runs of the base LLaVA architecture with variants of our method that freeze both the vision tower and LLM backbone. ASaRG also improves upon segmentation-assisted report generation models in the literature in terms of semantic information.

Furthermore, our proposed method can be easily extended or upgraded by exchanging the source models for the intermediate features or segmentation maps with more performant variants as they become available.

Finally, ASaRG lays the foundations for a new style of grounded report generation, as statements made in the generated report can be compared against the corresponding segmentation maps, allowing identification of obvious false positives and false negatives in some cases.

%% file: supplementary_v2.tex
\section*{Reproducibility Statement}
\label{sec: Reproducibility Statement}

For the purpose of reproduction, extension, or review, we refer to relevant code here:
\begin{itemize}
    \item{Training and evaluation code are available at \url{https://github.com/TIO-IKIM/ASaRG}.}
    \item{The CXAS codebase can be found at \url{https://github.com/ ConstantinSeibold/ChestXRayAnatomySegmentation}, although code or datasets for the additional classes are not included.}
    \item{The original LLaVA codebase, which the above codebase is built on top of, can be found at \url{https://github.com/haotian-liu/LLaVA}.}
\end{itemize}

\section*{Supplementary Materials}
\label{sec: Supplementary Materials}

\subsection*{Implementation Details}
\label{subsec: Implementation Details}

All experiments use the default Vicuna-7B LLM \cite{zheng_judging_2023} and ViT L/14 vision tower (336x336 resolution) pretrained on ImageNet \cite{deng_imagenet_2009} using CLIP \cite{radford_learning_2021}, with only the multi-modal adapter layer being modified or extended as described above. The pretrained LLM encoder of the regular LLaVA is chosen as the starting point over LLaVA-med weights because the former afforded the finetuning step on MIMIC-CXR data a greater degree of stability. In another, well-cited study \cite{hyland_maira-1_2024}, LLaVA-v1.5 also consistently outperformed LLaVA-med as a baseline. All experiments start on this pretrained baseline, and perform either a 1-epoch training run over all MIMIC-CXR training data or a two-stage training+finetuning run, after which an evaluation is performed on the holdout test set. The 1-/2-epoch limit is set to reduce the computational cost of experiments to manageable levels.

All experiments were conducted on a NVIDIA DGX node with 120 CPU cores, 4 NVIDIA A100 GPUs with 80GB VRAM each, and 800 GB of memory available. Hyperparameters were chosen by starting with default values from the official LLaVA repository's finetuning scripts, and then adapted to optimally use resources on our specific GPU node. For finetuning, ZeRO-2 offloading \cite{273920} is used, implemented via deepspeed \cite{rasley_deepspeed_2020}, analogously to LLaVA. Both bf16 and tf32 data formats are enabled, and data is automatically converted according to performance estimates. Gradient checkpointing is enabled. Training is performed with a batch size of 16, across 4 devices, for a total of 64, or 16 across 2 GPUs with 2 gradient accumulation steps for two-stage training. We adopt a maximum token sequence length of 2048 for the LLM inputs. We note that when segmentation maps are included, even when enabling gradient accumulation over many steps, a significant amount of model parameters have to be frozen to accomodate the segmentation maps on the GPUs, which is why we opted to freeze the entire backbone and limit ourselves to two gradient accumulation steps to achieve the original total batch size of 64 - With more compute budget, these parameters could, however, be unfrozen, which would almost certainly further improve performance by a significant margin. Training uses the Adam optimizer \cite{kingma_adam_2017}, a base learning rate of $\lambda= 2*10^{-5}$, no weight decay, and a cosine learning rate schedule with a warmup ratio of $\omega=0.03$. Model sharding is not performed.

During evaluation, a temperature of $\tau=0.2$ is applied, while top-P filtering and beam search are disabled, in order to keep reports factual. The maximum number of generated tokens, excluding, prompts, is limited to 1024. Evaluation is performed one at a time on a singular GPU.

\subsection*{Compute overhead and training times}

\begin{table*}[t]
    \centering
    \begin{tabularx}{\textwidth}{|C|C|C|C|C|}
        \hline
        \textbf{Method} & \textbf{Num Workers} & \textbf{Training Time} & \textbf{max. GPU (VRAM)} 
        \newline Overhead & max. RAM \newline Overhead \\
        \hline
        LLaVA, fully FT'd \newline(2 Epochs) & $4$ & $\sim14h$ & $4x\sim62.5GB$ & $\sim25.0GB$ \\
        \hline
        ASaRG, two-stage, +F (1+1 Epochs) & $4$ & $\sim7.2h\newline\sim5.2h$ & $4*\sim62.6GB\newline4*41.0-45.5GB$ & $\sim25.4GB$ \\
        \hline
        ASaRG, +F \newline(2 Ep.) & $4$ & $\sim14.5h$ & $4*\sim62.6GB$ & $\sim25.4GB$ \\
        \hline
        ASaRG, two-stage, +F+S (1+1 Ep.) & $2$ & $\sim7.2h\newline\sim14.5h$ & $4*\sim62.6GB\newline4*\sim54.3GB$ & $\sim523.0GB\newline\sim431.5-460.0GB^1$ \\
        \hline
        ASaRG, two-stage, +F+SCoS (1+1 Ep.) & $2$ & $\sim7.2h\newline\sim14.7h^2$ & $4*\sim62.6GB\newline4x\sim47.0GB$ & $\sim523.0GB\newline\sim431.5-460.0GB^1$ \\
        \hline
    \end{tabularx}
    \caption{\textbf{Compute overhead comparison for tested settings}
    \small
    +F denotes added image features, +S denotes added segmentation maps, +SCoS denotes added superclass-only segmentation maps. Peak RAM usage refers to training with MIMIC-CXR plus associated segmentation maps and will vary wildly depending on dataset size. 
    Note that Linux will often autonomously cache data from the disk, which means startup or execution are sometimes randomly faster depending on the available system RAM which Linux can cache into and cache contents. \\
    $^1$ Peak RAM usage is significantly higher than consistent RAM usage. This peak occurs at the start of training, when the segmentation maps are cached, and is only experienced briefly. The second number denotes consistent RAM usage. \\
    $^2$ Note that when superclasses are aggregated as preprocessing, instead of during training, their RAM impact reduces by a factor of 10. Doing so, although not done during our experiments, allows efficiently training with at least 4 workers and no gradient accumulation again, improving the training time from 7+14.5 hours back to 7+7.5 hours. In this scenario, peak and consistent RAM usage would both decrease by a factor slightly lower than 10, as the segmentation maps make up the bulk of RAM usage. Alternatively, when crossing a break point of just above 1TB RAM, 4 workers can be used without extra preprocessing or other changes.}
    \label{tab: Compute overhead comparison}
\end{table*}

\begin{table*}[t]
    \centering
    \begin{tabularx}{\textwidth}{|C|C|C|C|}
        \hline
        \textbf{Source} & \textbf{ASaRG} & \textbf{LLaVA} & \textbf{Ground Truth} \\
        \hline
        Avg. ``Human-ness'' & 0.14 & 0.48 & 0.85 \\
        \hline
    \end{tabularx}
    \caption{\textbf{Style assessment} - This table shows average style rating across 100 samples, on a Likert scale from -2 (almost certainly AI-generated) to 2 (almost certainly human-made). Uncertainty on these averages is $\sigma\approx0.1$ if estimated as a counting error.}
    \label{tab: lexical quality control}
\end{table*}

We note that some of the compute constraints because of which the LLaVA backbone is frozen when segmentations are introduced can be alleviated without the use of additional GPU resources. Runtime is bottlenecked in most practical scenarios by the memory overhead of the object holding the compressed segmentation maps, and therefore by number of dataset workers. A breakpoint of slightly above 1TB peak memory usage allows 4 workers for runtimes comparable to baseline LLaVA. Consequently, the experiments can be up-scaled with larger LLMs or without freezing any model parameters rather easily, e.g. on a DGX-2 node or a cluster on which all compute resources are available without node-specific limitations, but only efficiently if the data loading can be up-scaled. During deployment/inference, only one GPU is required, and peak RAM overhead is limited to the original LLaVA overhead, plus a comparatively negligible amount (<1GB) for the segmentation maps of the input image.

\subsection*{Converting MIMIC-CXR to a VQA format}
\label{subsec: Converting MIMIC-CXR to a VQA format}

We convert the MIMIC-CXR dataset \cite{johnson_mimic-cxr_2019} into a VQA format by extracting radiological reports $r_i$ with images $v_i$. For each report $r_i$, where unambiguously possible and extant, we extract the report's findings section $r_i^F$ and generate single-turn chats $\{user: v_i u_i^F, assistant: r_i^F\}_{X\in\{F, I\}}$ with user queries $u_i^F \in P_F $ being one of a diverse set of 33 paraphrasations $P_F$ (cf.~\ref{fig:instructions-findings}) querying for findings from the image $v_i$. Dataset splits otherwise follow the official MIMIC-CXR splits.

Additionally, we have a clinical expert write multi-turn chats for 206 radiology reports $r_i$ with images $v_i$, mimicking real clinician-model interactions that may involve both the findings and impressions sections. We use these to fine-tune a GPT 3.5 model \cite{NEURIPS2020_1457c0d6} for clinical visual instruction generation and generate another 5,000 open-ended multi-turn chats from MIMIC-CXR data, which are added to the training data, in order to improve training stability and generalizability of our finetuned models.

\begin{figure*}[t]
\centering
\begin{tcolorbox}[colback=gray!5, colframe=gray!50!black]
\begin{enumerate}[leftmargin=*, itemsep=0mm]
\item Can you describe what you see in the image?
\item Please provide an overview of the key observations in the X-ray images.
\item What are the significant details captured in this medical image?
\item Summarize the visual findings from this medical scan.
\item Give me a brief summary of the image's diagnostic features.
\item Can you outline the main points of interest in this picture?
\item What observations can be made from this radiological image?
\item Describe the significant findings in this visual information.
\item Offer a brief overview of the diagnostic details in the picture.
\item List the main points of interest in this radiological data.
\item Outline the relevant findings of this medical imaging.
\item Give me a summarized account of the observations here.
\item Provide a concise summary of the diagnostic features.
\item Can you identify the key takeaways from this visual data?
\item Highlight the significant findings in this X-ray image.
\item Summarize the important aspects of this radiological data.
\item Offer a brief synopsis of the observations captured.
\item Describe the most salient features in this X-ray image.
\item What do you perceive as the primary diagnostic insights from this picture?
\item Provide details about any notable and unremarkable features in the image.
\item Describe the overall condition of the subject in the image.
\item Can you summarize the key observations from this radiograph?
\item Discuss the significant findings within this X-ray image.
\item Brief me on the findings.
\item What can you see on the X-ray images?
\item Please provide a summary of the observations made in the images, noting any abnormalities or potential issues.
\item Describe what you see in the images and mention if any areas appear normal or unremarkable.
\item Summarize the key observations and abnormalities that stand out in the images.
\item Give an overview of the findings.
\item Summarize the overall impression of the images, emphasizing critical observations.
\item Provide a concise summary of the findings using medical jargon.
\item Are there any notable or unremarkable findings that should be known to the patient's primary care physician?
\item Summarize the findings in a manner that allows for easy communication with the patient's healthcare team.
\end{enumerate}
\end{tcolorbox}
\caption{\textbf{User queries} - The user queries $u_i^F \in P_F$ for prompting a VLM for radiology findings on an image $v_i$, each paired with the findings $r_i^F$ as the response to form a single-turn chat.}
\label{fig:instructions-findings}
\end{figure*}

\subsection*{Lexical Quality Control}

To reaffirm that ASaRG produces sound reports despite low lexical metric scores - as indicted by the CE metric performance - we conducted a qualitative survey of 300 sample reports. 100 reports were each drawn from the highest-performing ASaRG, the fully finetuned LLaVA baseline, and ground truths. All reports were randomly selected from the MIMIC-CXR test set, and the assessing clinician was blinded with respect to their numbers and origin. The survey included three questions for each report. Firstly, whether the report is grammatically sound. Secondly, whether the report is clinically sensible in principle, i.e. containing no false abbreviations or combination of observations impossible in one individual. Finally, the clinician rated the reports on a Likert scale from -2 to 2, denoting that a report is \textit{almost certainly AI-generated} (-2), \textit{probably AI-generated} (-1), \textit{uncertain} (0), \textit{probably human-made} (1), \textit{almost certainly human-made} (2). To not conflate this assessment with report preference or factuality, associated images were not provided, only the reports themselves. \\
Across all samples in the survey, the clinician did not find any examples containing grammatically nonsensical or clinically unlikely/impossible reports, strongly implying that ASaRG is well-capable of generating believable and accurate reports in our VQA setting, as the CE metrics suggested. Notably, both ASaRG and the LLaVA baseline appear to have developed a style somewhat distinct from human reports and one another. The distinction from human reports is likely explained with the conversational style of the VQA setting and our additional multi-turn conversation data, although where the ASaRG-LLaVA distinction comes from is unclear. Interestingly, even with this distinction, the average score for both methods is still above 0, indicating that generated reports are still often pass for human. Similarly, for both methods, examples are found for the entire scale, meaning that no trends are completely universal. \\
It was noted that a few distinguishing features of generated reports stood out compared to the ground truth. Firstly, generated reports have a tendency to be written like sentences in a conversation, probably due to the VQA training setting, often containing phrases such as ``There is X seen in [...]'', where clinicians often prefer abbreviated styles such as ``X noted.''. Furthermore, generated reports often contained mentions of related tissue regions or findings in unrelated parts of the report. While not producing incorrect results, clinicians typically impose a structure on the report, related either to the specific indication or by clustering associated regions/findings in the body and in the report. Finally, we note that generated reports typically, though not always, provide comparatively stronger certainty on results. Where a clinician may report ``Observation in region X may be evidence of pathology Y.'', the generated report may contain the sentence ``Pathology Y observed in region X.''

\subsection*{ASaRG performance gains are not class-specific}
\label{subsec: ASaRG Performance gains are not class-specific}

\begin{figure*}[t]
    \centering
    \includegraphics[width=0.95\linewidth]{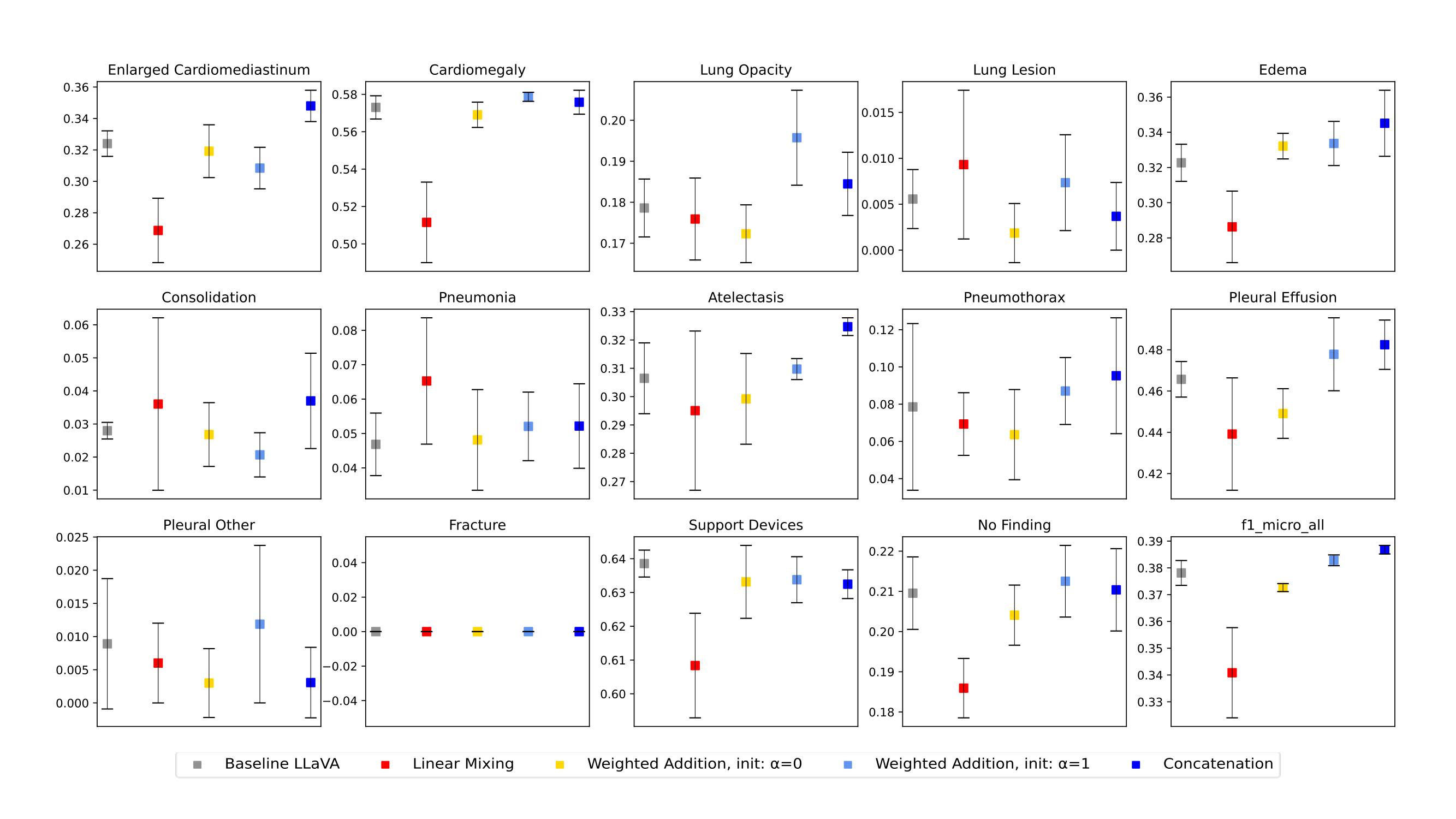}
    \caption{\textbf{Per-class performance} - This series of plots shows the per-class performance of different variants of ASaRG that use additional intermediate features, as well as the accompanying baselines.}
    \label{fig: Per-class performance Set 1}
\end{figure*}
\begin{figure*}[t]
    \centering
    \includegraphics[width=0.95\linewidth]{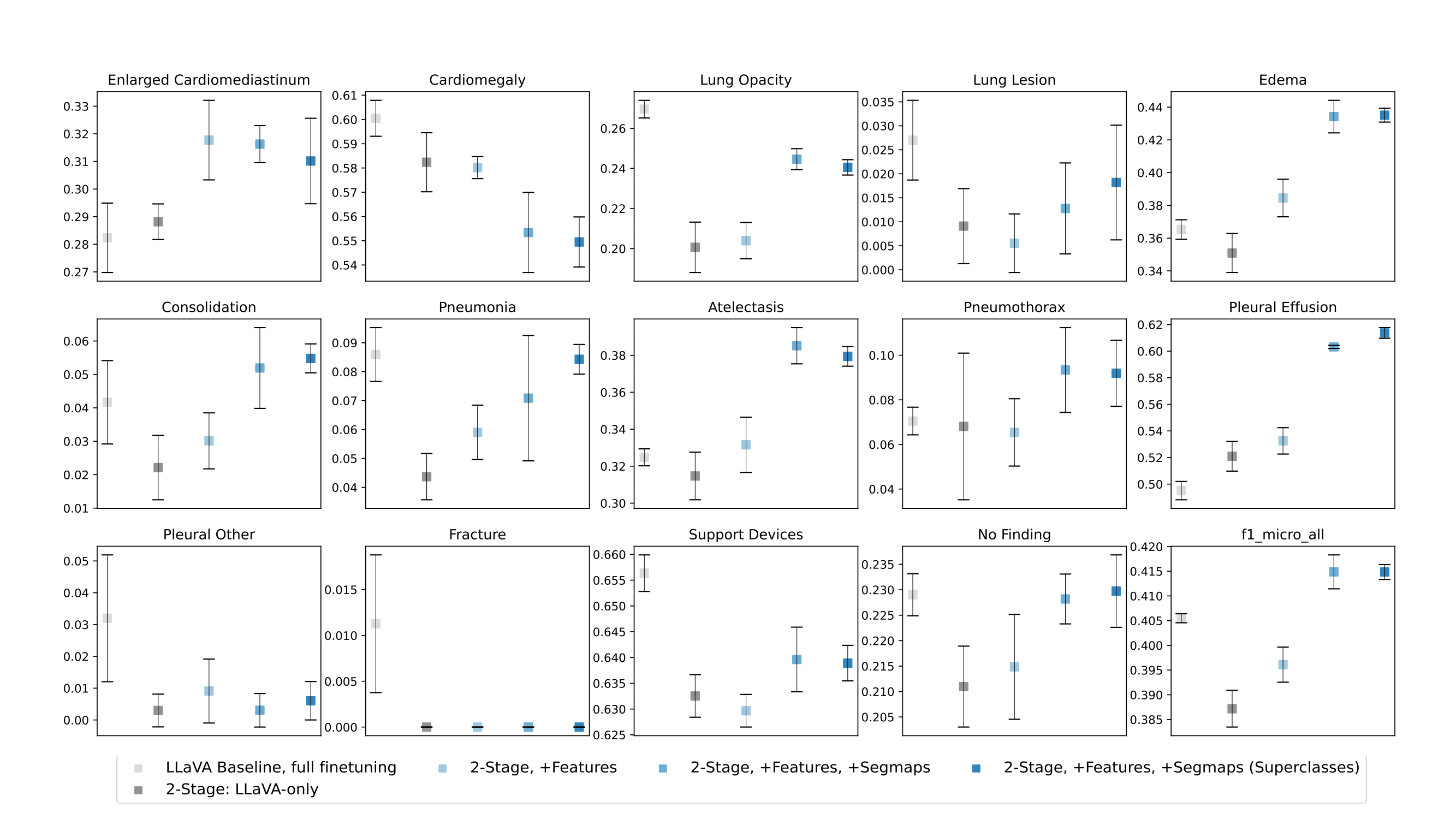}
    \caption{\textbf{Per-class performance} - This series of plots shows the per-class performance of different variants of ASaRG that use additional intermediate features and (fine-grained) segmentation maps, as well as the accompanying baselines.}
    \label{fig: Per-class performance Set 2}
\end{figure*}

Figs. \ref{fig: Per-class performance Set 1} and \ref{fig: Per-class performance Set 2} display the per-class performance of different versions of ASaRG. It appears that while ASaRG confers some measure of advantage to the report generation model, the exact nature of this advantage is surprisingly diffuse and does not relate to a specific class. When the same experiment is repeated, we found that while the magnitude of the advantage remained similar, the specific classes in which an advantage or even disadvantage was exhibited did not always. When comparing to other work which reports per-class performances, e.g. \cite{hyland_maira-1_2024}, we observe that ASaRG seems to significantly underperform or overperform in some classes. While a general performance gap and behavior change is expected in our experiments - since we only train for 1 or 2 epochs at a time, freeze parameters, and work in a VQA setting - we have thus far not found a good explanation for the class-specific differences to literature that we observe. 
Finally, we note that due to the limited nature of the experiments and class imbalances in the MIMIC-CXR test set, the performance difference on some classes, such as fractures, could not be determined.

\subsection*{Failure Mode Analysis: Examples}
\label{Failure Mode Analysis: Examples}

The following pages contain some randomly chosen examples for failure modes discussed in the Discussion section. Each example contains the associated image, ground truth report, and the report generated by the highest-scoring ASaRG checkpoint. While they nicely illustrate some common concerns, other, less approachable failure modes that simply relate to the black-box nature of neural networks remain undiscussed, but should remain in the mind of any researchers attempting to bring medical report generation into the clinical practice. Note that we truncated ground truths and generated reports to pertinent sections for readability. \\

\clearpage

\begin{figure*}[t]
    \centering
    \includegraphics[width=\linewidth]{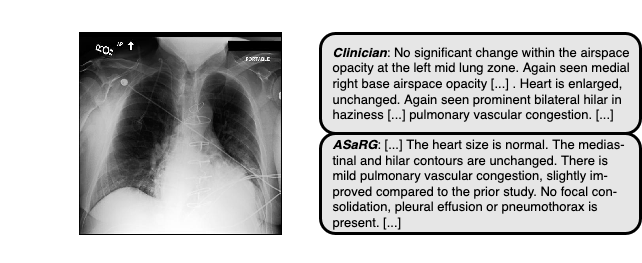}
    \caption{\textbf{Failure Mode: Prior Hallucinations} - ASaRG and LLaVA, across variants, have a tendency to refer to prior reports. This behavior is likely learned because a significant fraction of ground truth reports reference prior reports. Whether part of MIMIC-CXR or not, such prior reports are not available in the style of one-report-one-image report generation ASaRG is trained on, and information presented in that context is typically also hallucinated.}
    \label{fig: Failure Mode Example 1}
\end{figure*}

\begin{figure*}[t]
    \centering
    \includegraphics[width=\linewidth]{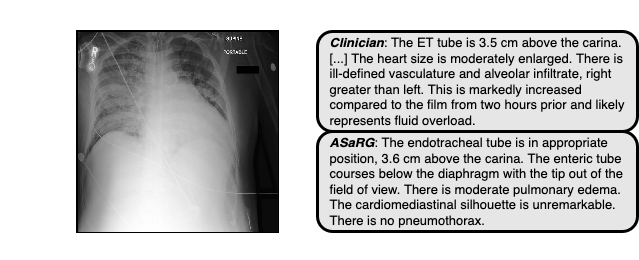}
    \caption{\textbf{Failure Mode: Length Hallucinations} - ASaRG and LLaVA, across variants, report sizes or distances in centimeters. This typically happens in the presence of clearly visible ET tubes, but is observed for nodules and other objects as well. Generally, these numbers are hallucinated, likely even in examples where they are correct (such an example is depicted here), as ASaRG lacks a reference against which to measure lengths, even if it can make educated guesses based on learned human anatomy.}
    \label{fig: Failure Mode Example 2}
\end{figure*}

\begin{figure*}[t]
    \centering
    \includegraphics[width=\linewidth]{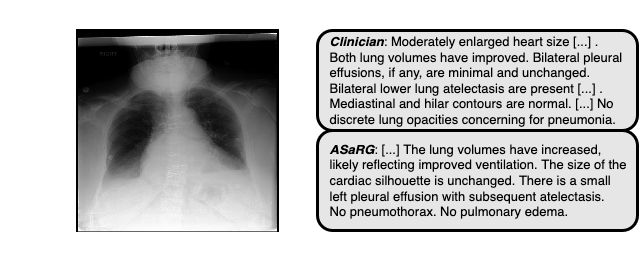}
    \caption{\textbf{Failure Mode: Valid Reports} - In some cases, ASaRG presents valid reports, which are different from the ground truth and then mischaracterized by metrics, proving that even CE metrics are not infallible. One such example is depicted here.}
    \label{fig: Failure Mode Example 4}
\end{figure*}

\begin{figure*}[t]
    \centering
    \includegraphics[width=\linewidth]{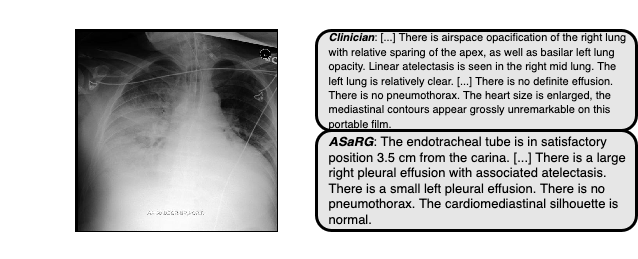}
    \caption{\textbf{Failure Mode: Overconfidence} - ASaRG and LLaVA are particularly prone to presenting findings as overlay certain. An example is again depicted here, where clinicians may prefer to be less explicit about hard facts, and instead frame findings as potential evidence. Note that neither overconfidence in generated reports or moderation on behalf of radiologists is absolute, and plentiful counterexamples exist.}
    \label{fig: Failure Mode Example 5}
\end{figure*}

\begin{figure*}[t]
    \centering
    \includegraphics[width=\linewidth]{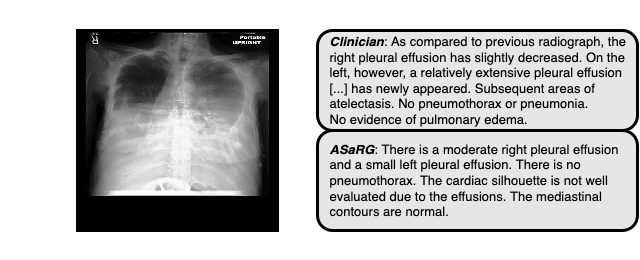}
    \caption{\textbf{Failure Mode: Under-reporting} - ASaRG and LLaVA systematically under-report pathologies. This can result in false negatives, which are common, and partially false reports, where only one or several, but not all pathologies are correctly identified. This figure exhibits and example of an otherwise correct report missing a key diagnosis, atelectasis. Note that in some cases, MIMIC-CXR labels aren't absolute truths and clinicians may disagree about diagnoses, meaning not \textit{every} report missing one of several pathologies is necessarily false.}
    \label{fig: Failure Mode Example 6}
\end{figure*}

\begin{figure*}[t]
    \centering
    \includegraphics[width=\linewidth]{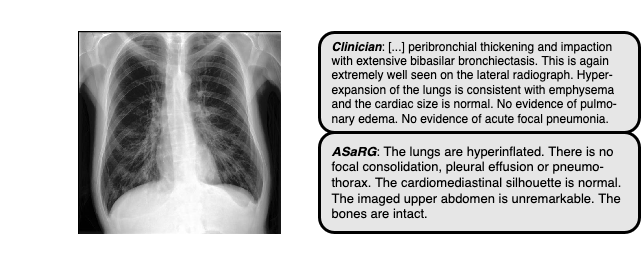}
    \caption{\textbf{Failure Mode: Missing Views} - Missing views, similar to missing prior reports, represent a systemic issue with training in a one-to-one fashion. In this showcased example, a lateral view is referenced as a source in the ground truth, but said lateral view is not available to the model to draw conclusions from. Training a model with multiple views or subsequent images in one data point has recently been shown as an effective mitigation strategy \cite{bannur_maira-2_2024}}.
    \label{fig: Failure Mode Example 7}
\end{figure*}

\clearpage

\begin{table*}[!h]
    \centering
    \begin{tabularx}{\textwidth}{|>{\arraybackslash}p{1cm}|X|>{\arraybackslash}p{1cm}|X|}
        \hline
        \textbf{\#} & \textbf{Class name} & \textbf{\#} & \textbf{Class name} \\
        \hline
        0 & "spine" & 36 & "posterior 12th rib right" \\
        \hline
        1 & "cervical spine" & 37 & "posterior 12th rib left" \\
        \hline
        2 & "thoracic spine" & 38 & "anterior 11th rib right" \\
        \hline
        3 & "lumbar spine" & 39 & "posterior 11th rib right" \\
        \hline
        4 & "vertebrae C1" & 40 & "anterior 11th rib left" \\
        \hline
        5 & "vertebrae C2" & 41 & "posterior 11th rib left" \\
        \hline
        6 & "vertebrae C3" & 42 & "anterior 10th rib right" \\
        \hline
        7 & "vertebrae C4" & 43 & "posterior 10th rib right" \\
        \hline
        8 & "vertebrae C5" & 44 & "anterior 10th rib left" \\
        \hline
        9 & "vertebrae C6" & 45 & "posterior 10th rib left" \\
        \hline
        10 & "vertebrae C7" & 46 & "anterior 9th rib right" \\
        \hline
        11 & "vertebrae T1" & 47 & "posterior 9th rib right" \\
        \hline
        12 & "vertebrae T2" & 48 & "anterior 9th rib left" \\
        \hline
        13 & "vertebrae T3" & 49 & "posterior 9th rib left" \\
        \hline
        14 & "vertebrae T4" & 50 & "anterior 8th rib right" \\
        \hline
        15 & "vertebrae T5" & 51 & "posterior 8th rib right" \\
        \hline
        16 & "vertebrae T6" & 52 & "anterior 8th rib left" \\
        \hline
        17 & "vertebrae T7" & 53 & "posterior 8th rib left" \\
        \hline
        18 & "vertebrae T8" & 54 & "anterior 7th rib right" \\
        \hline
        19 & "vertebrae T9" & 55 & "posterior 7th rib right" \\
        \hline
        20 & "vertebrae T10" & 56 & "anterior 7th rib left" \\
        \hline
        21 & "vertebrae T11" & 57 & "posterior 7th rib left" \\
        \hline
        22 & "vertebrae T12" & 58 & "anterior 6th rib right" \\
        \hline
        23 & "vertebrae L1" & 59 & "posterior 6th rib right" \\
        \hline
        24 & "vertebrae L2" & 60 & "anterior 6th rib left" \\
        \hline
        25 & "vertebrae L3" & 61 & "posterior 6th rib left" \\
        \hline
        26 & "vertebrae L4" & 62 & "anterior 5th rib right" \\
        \hline
        27 & "vertebrae L5" & 63 & "posterior 5th rib right" \\
        \hline
        28 & "rib\_cartilage" & 64 & "anterior 5th rib left" \\
        \hline
        29 & "sternum" & 65 & "posterior 5th rib left" \\
        \hline
        30 & "clavicles" & 66 & "anterior 4th rib right" \\
        \hline
        31 & "clavicle left" & 67 & "posterior 4th rib right" \\
        \hline
        32 & "clavicle right" & 68 & "anterior 4th rib left" \\
        \hline
        33 & "scapulas" & 69 & "posterior 4th rib left" \\
        \hline
        34 & "scapula left" & 70 & "anterior 3rd rib right" \\
        \hline
        35 & "scapula right" & 71 & "posterior 3rd rib right" \\
        \hline
    \end{tabularx}
    \caption{\textbf{The ASaRG Segmentation Classes (1/3)}}
    \label{tab: ASaRG Segmentation Classes 1}
\end{table*}

\begin{table*}[!h]
    \centering
    \begin{tabularx}{\textwidth}{|>{\arraybackslash}p{1cm}|X|>{\arraybackslash}p{1cm}|X|}
        \hline
        \textbf{\#} & \textbf{Class name} & \textbf{\#} & \textbf{Class name} \\
        \hline
        72 & "anterior 3rd rib left" & 108 & "stomach" \\
        \hline
        73 & "posterior 3rd rib left" & 109 & "small bowel" \\
        \hline
        74 & "anterior 2nd rib right" & 110 & "duodenum" \\
        \hline
        75 & "posterior 2nd rib right" & 111 & "liver" \\
        \hline
        76 & "anterior 2nd rib left" & 112 & "pancreas" \\
        \hline
        77 & "posterior 2nd rib left" & 113 & "kidney left" \\
        \hline
        78 & "anterior 1st rib right" & 114 & "kidney right" \\
        \hline
        79 & "posterior 1st rib right" & 115 & "cardiomediastinum" \\
        \hline
        80 & "anterior 1st rib left" & 116 & "upper mediastinum" \\
        \hline
        81 & "posterior 1st rib left" & 117 & "lower mediastinum" \\
        \hline
        82 & "12th rib" & 118 & "anterior mediastinum" \\
        \hline
        83 & "posterior 11th rib" & 119 & "middle mediastinum" \\
        \hline
        84 & "anterior 11th rib" & 120 & "posterior mediastinum" \\
        \hline
        85 & "posterior 10th rib" & 121 & "heart" \\
        \hline
        86 & "anterior 10th rib" & 122 & "heart atrium left" \\
        \hline
        87 & "posterior 9th rib" & 123 & "heart atrium right" \\
        \hline
        88 & "anterior 9th rib" & 124 & "heart myocardium" \\
        \hline
        89 & "posterior 8th rib" & 125 & "heart ventricle left" \\
        \hline
        90 & "anterior 8th rib" & 126 & "heart ventricle right" \\
        \hline
        91 & "posterior 7th rib" & 127 & "aorta" \\
        \hline
        92 & "anterior 7th rib" & 128 & "ascending aorta" \\
        \hline
        93 & "posterior 6th rib" & 129 & "descending aorta" \\
        \hline
        94 & "anterior 6th rib" & 130 & "aortic arch" \\
        \hline
        95 & "posterior 5th rib" & 131 & "pulmonary artery" \\
        \hline
        96 & "anterior 5th rib" & 132 & "pulmonary trunc" \\
        \hline
        97 & "posterior 4th rib" & 133 & "left pulmonary artery" \\
        \hline
        98 & "anterior 4th rib" & 134 & "right pulmonary artery" \\
        \hline
        99 & "posterior 3rd rib" & 135 & "inferior vena cava" \\
        \hline
        100 & "anterior 3rd rib" & 136 & "esophagus" \\
        \hline
        101 & "posterior 2nd rib" & 137 & "lung" \\
        \hline
        102 & "anterior 2nd rib" & 138 & "right lung" \\
        \hline
        103 & "posterior 1st rib" & 139 & "left lung" \\
        \hline
        104 & "anterior 1st rib" & 140 & "lung base" \\
        \hline
        105 & "diaphragm" & 141 & "mid zone lung" \\
        \hline
        106 & "left hemidiaphragm" & 142 & "upper zone lung" \\
        \hline
        107 & "right hemidiaphragm" & 143 & "apical zone lung" \\
        \hline
    \end{tabularx}
    \caption{\textbf{The ASaRG Segmentation Classes (2/3)}}
    \label{tab: ASaRG Segmentation Classes 2}
\end{table*}

\begin{table*}[!h]
    \centering
    \begin{tabularx}{\textwidth}{|>{\arraybackslash}p{1cm}|X|>{\arraybackslash}p{1cm}|X|}
        \hline
        \textbf{\#} & \textbf{Class name} & \textbf{\#} & \textbf{Class name} \\
        \hline
        144 & "right upper zone lung & 180 & "Consolidation" \\
        \hline
        145 & "right mid zone lung & 181 & "Diffuse Nodule" \\
        \hline
        146 & "right lung base" & 182 & "Effusion" \\
        \hline
        147 & "right apical zone lung" & 183 & "Emphysema" \\
        \hline
        148 & "left upper zone lung" & 184 & "Fibrosis" \\
        \hline
        149 & "left mid zone lung" & 185 & "Fracture" \\
        \hline
        150 & "left lung base" & 186 & "Mass" \\
        \hline
        151 & "left apical zone lung" & 187 & "Nodule" \\
        \hline
        152 & "lung lower lobe left" & 188 & "Pleural Thickening" \\
        \hline
        153 & "lung upper lobe left" & 189 & "Pneumothorax" \\
        \hline
        154 & "lung lower lobe right" & 190 & "CVC - Normal" \\
        \hline
        155 & "lung middle lobe right" & 191 & "CVC - Borderline" \\
        \hline
        156 & "lung upper lobe right" & 192 & "NGT - Normal" \\
        \hline
        157 & "less obstructed lung" & 193 & "ETT - Normal" \\
        \hline
        158 & "l. obs. right lung" & 194 & "NGT - Incompletely Imaged" \\
        \hline
        159 & "l. obs left lung" & 195 & "CVC - Abnormal" \\
        \hline
        160 & "l. obs. lung base" & 196 & "ETT - Borderline" \\
        \hline
        161 & "l. obs. mid zone lung" & 197 & "Swan Ganz Catheter Present" \\
        \hline
        162 & "l. obs. upper zone lung" & 198 & "NGT - Borderline" \\
        \hline
        163 & "l. obs. apical zone lung" & 199 & "NGT - Abnormal" \\
        \hline
        164 & "l. obs. right upper zone lung" & 200 & "ETT - Abnormal" \\
        \hline
        165 & "l. obs. right mid zone lung" & 201 & "Fremdkörper" \\
        \hline
        166 & "l. obs. right lung base" & 202 & "Cable" \\
        \hline
        167 & "l. obs. right apical zone lung" & 203 & "Clamp" \\
        \hline
        168 & "l. obs. left upper zone lung" & 204 & "Electronics" \\
        \hline
        169 & "l. obs. left mid zone lung" & 205 & "Throat Pipe" \\
        \hline
        170 & "l. obs. left lung base" & 206 & "Pipe" \\
        \hline
        171 & "l. obs. left apical zone lung" & 207 & "Letters" \\
        \hline
        172 & "trachea" & 208 & "Stitches" \\
        \hline
        173 & "tracheal bifurcation" & 209 & "Sensors" \\
        \hline
        174 & "breast" & 210 & "Implant" \\
        \hline
        175 & "breast left" & 211 & "Foreign Object" \\
        \hline
        176 & "breast right" &  &  \\
        \hline
        177 & "Atelectasis" &  &  \\
        \hline
        178 & "Calcification" &  &  \\
        \hline
        179 & "Cardiomegaly" &  &  \\
        \hline
    \end{tabularx}
    \caption{\textbf{The ASaRG Segmentation Classes (3/3)}}
    \label{tab: ASaRG Segmentation Classes 3}
\end{table*}